\documentclass{article}

    \PassOptionsToPackage{numbers}{natbib}

\usepackage[main, nonanonymous]{neurips_2026}
\makeatletter
\renewcommand{\@notice}{}
\makeatother
\AtBeginDocument{\nolinenumbers}

\usepackage{times}
\usepackage{latexsym}
\usepackage{graphicx}
\usepackage{subcaption} 
\usepackage[T1]{fontenc}
\usepackage{multirow}

\usepackage[utf8]{inputenc}
\usepackage{booktabs}
\usepackage{siunitx} 
\usepackage{microtype}

\usepackage{inconsolata}
\usepackage{booktabs}  
\usepackage{array}     
\usepackage[table]{xcolor} 
\usepackage{wrapfig}

\usepackage{graphicx}
\usepackage{tabularx}
\usepackage{booktabs}
\usepackage{multirow}
\usepackage{graphicx}
\usepackage{amsmath}

\usepackage{pgfplots}
\pgfplotsset{compat=1.18}

\definecolor{myblue}{RGB}{79, 129, 189}    
\definecolor{myorange}{RGB}{192, 80, 77}   
\definecolor{gridgray}{RGB}{220, 220, 220} 

\usepackage[utf8]{inputenc} 
\usepackage[T1]{fontenc}    
\usepackage{hyperref}       
\hypersetup{
    colorlinks=true,
    citecolor=[rgb]{0, 0.4, 0.7}, 
    linkcolor=red,               
    urlcolor=blue                
}
\usepackage{url}            
\usepackage{booktabs}       
\usepackage{amsfonts}       
\usepackage{nicefrac}       
\usepackage{microtype}      
\usepackage{xcolor}         

\usepackage{enumitem}

\title{BlockPilot: Instance-Adaptive Policy Learning for Diffusion-based Speculative Decoding}

\author{
    Hao Zhang \quad Yiming Hu\footnotemark[2] \quad Yong Wang\footnotemark[2] \hspace{0.8em} 
    Mingqiao Mo \quad Xin Xiao \quad Xiangxiang Chu \\[1.2ex]
    AMAP, Alibaba Group \\
    \normalsize\href{https://github.com/AMAP-ML/BlockPilot}{https://github.com/AMAP-ML/BlockPilot}
}

\begin{document}

\maketitle

\begin{abstract}
Speculative decoding accelerates inference by using a lightweight draft model to generate candidate tokens in parallel, and are then verified by the target model, enabling lossless acceleration. Recently, diffusion-based speculative decoding further improves parallelism by generating multiple tokens per forward pass via block-level diffusion, achieving state-of-the-art (SOTA) performance. However, existing methods adopt a fixed inference block size and assume a uniform optimal decoding strategy across all inputs. In this paper, we show that this assumption is suboptimal, as the optimal block size varies across samples and plays a critical role in speculative decoding performance. Moreover, these values exhibit a clear local structure, concentrating around the training block size, which reduces the problem to a low-dimensional and structured decision space. Based on these insights, we propose BlockPilot, a sample-adaptive policy that predicts the optimal block size from the prefilling representation. Specifically, we formulate block size selection as a lightweight policy learning problem and propose an instance-adaptive decision mechanism that predicts the optimal block size based on the representation of the prefilling stage. The prediction is performed only once after prefilling, allowing for seamless integration. Extensive experiments demonstrate that our method is plug-and-play, introduces minimal overhead, and consistently improves efficiency, achieving an acceptance length of 5.92 and a 4.20$\times$ speedup on Qwen3-4B under temperature $T=1$.

{
\renewcommand{\thefootnote}{\fnsymbol{footnote}}
\footnotetext[2]{\ Project lead and corresponding author.}
}

\end{abstract}

\section{Introduction}
\begin{wrapfigure}{r}{0.46\textwidth}
\vspace{-10pt}
\centering
\includegraphics[width=0.46\textwidth]{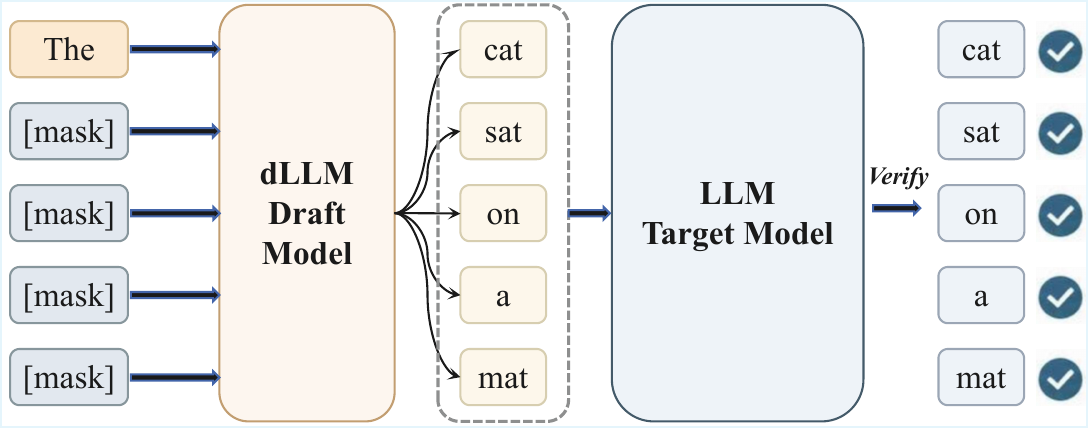}
\vspace{-8pt}
\caption{Diffusion-based speculative decoding with a dLLM draft model. The dLLM proposes a block of tokens in parallel, while the target LLM verifies the block and accepts the longest consistent prefix.}
\label{fig:sp}
\vspace{-10pt}
\end{wrapfigure}

Large Language Models (LLMs) \cite{touvron2023open,yang2025qwen3,chiang2023vicuna} have achieved remarkable performance across a wide range of tasks \citep{annepaka2025large,hadi2023large}, demonstrating strong capabilities in reasoning, code generation, and open-ended dialogue. Despite these advances, their inference efficiency is still fundamentally constrained by token-by-token autoregressive decoding \cite{tay2022efficient,papa2024survey,wan2023efficient}. Since each token must be generated conditioned on previously produced tokens, the decoding process is inherently sequential, leading to high latency and limited parallelism, especially for long-form generation. To alleviate this bottleneck, speculative decoding \cite{leviathan2023fastinferencetransformersspeculative, li2025eaglespeculativesamplingrequires, li2024eagle2fasterinferencelanguage, li2025eagle3scalinginferenceacceleration, cai2024medusasimplellminference} introduces a lightweight draft model to propose multiple candidate tokens ahead of the target model. These candidates are then verified by the target model in parallel, allowing multiple tokens to be accepted within a single decoding step. As a result, speculative decoding enables lossless acceleration without altering the output distribution of the target model.


In recent years, diffusion-based language models (dLLMs) \cite{nie2025largelanguagediffusionmodels, arriola2025blockdiffusioninterpolatingautoregressive, wu2025fastdllmv2efficientblockdiffusion, cheng2025sdarsynergisticdiffusionautoregressionparadigm} have emerged as a promising direction, and diffusion-driven speculative decoding further improves parallelism. By using a dLLM as the draft model, block-wise diffusion enables the generation of multiple tokens in a single forward pass, significantly reducing decoding latency and improving hardware utilization. Fig.~\ref{fig:sp} illustrates the paradigm of speculative decoding based on diffusion models. However, early approaches \cite{li2025diffuspecunlockingdiffusionlanguage,sandler2025specdiff2scalingdiffusiondrafter,samragh2025llmknowsfutureuncovering} rely on large diffusion models and exhibit weak coupling with the target model, limiting their practicality. More recently, DFlash~\cite{chen2026dflash} achieves the first practically deployable state-of-the-art (SOTA) performance in block diffusion-based speculative decoding. By injecting hidden representations from the target model into the diffusion draft model, it enables high-quality parallel drafting, substantially improving acceptance length and real-world speedup. 


Although this paradigm offers strong potential for parallelism, existing methods typically use a fixed block size during inference, directly inherited from the training stage. This design is simple and easy to deploy; however, it overlooks a critical dimension: the decoding policy. Existing methods assume that a single block size is optimal for all inputs and treat it as a static hyperparameter. We argue that this assumption is fundamentally suboptimal. The optimal degree of parallelism depends on input-specific predictability, making block size a sample-dependent decision. Inputs differ in semantic constraints, contextual determinism, and token-level predictability, which lead to varying tolerance for parallel drafting. While larger block sizes improve efficiency for constrained continuations, they may cause error accumulation and lower acceptance for less predictable trajectories; smaller block sizes are more conservative but may underutilize the parallel capacity of diffusion-based generation. Therefore, a fixed block size policy is misaligned with input-level generation characteristics and leaves potential acceleration gains unexplored.

\begin{figure}[t]
    \centering
    \includegraphics[width=0.95\textwidth]{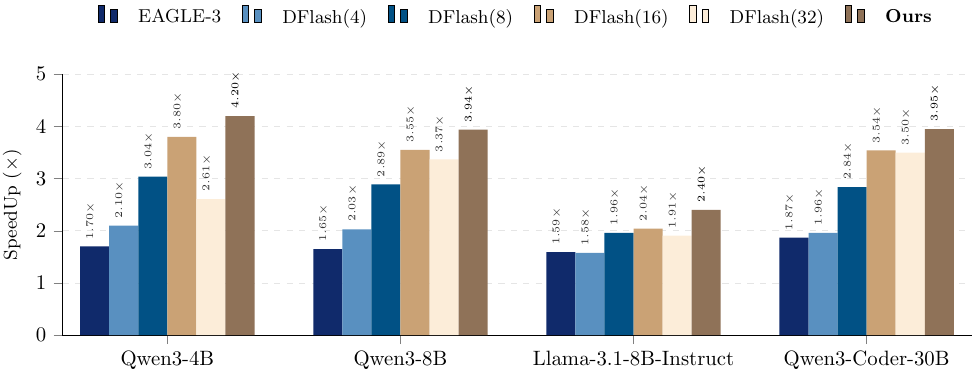}
    \caption{Speedup comparison across models under temperature $T=1$. Our method achieves the highest acceleration across all settings. Here, DFlash$(n)$ denotes DFlash with block size $n$.}
    \label{fig:performance_comparision}
\end{figure}

In this paper, we revisit diffusion-based speculative decoding from a largely overlooked perspective. Beyond designing stronger draft models or more efficient verification mechanisms, we ask whether the decoding strategy itself should be treated as a learnable component. We argue that block size is not merely an engineering parameter, but a key control variable that determines the acceptance length in speculative decoding. To validate this insight, we conduct a systematic block size sweep across multiple representative datasets. The results show a clear mismatch between sample-wise optimal block sizes and the fixed configuration used during training. Only a subset of samples achieve optimal performance under the default setting, while many prefer different inference-time block sizes. This suggests that existing fixed strategies fail to fully exploit the efficiency potential of diffusion-based speculative decoding.

Furthermore, we observe that although the optimal block size varies across samples, its distribution is not arbitrarily scattered. Instead, it exhibits a clear local structure: for most samples, the optimal block size concentrates within a narrow region around the training configuration, and few samples achieve optimal performance outside this region. This locality has an important methodological implication. It transforms what would otherwise require expensive online search into a small-scale, discrete, and well-structured classification problem. In other words, sample-adaptive block size selection does not require complex dynamic optimization, and can instead be efficiently handled by a lightweight predictor.

Based on this insight, we introduce BlockPilot, a sample-adaptive block size selection method. Specifically, after the target model completes prefilling, we use the predictive distribution of the final token as a representation of the current decoding state. Since this token has aggregated full-context information via autoregressive attention, its distribution reflects not only local uncertainty but also the model’s estimation of future generation stability, serving as an effective signal for predicting the acceptable block length. Therefore, we formulate block size selection as a policy learning problem over a discrete local action space and approximate the optimal policy with a lightweight predictor. The prediction is performed only once after prefilling and does not modify the target model, draft model, or verification process, allowing seamless integration into existing diffusion-based speculative decoding frameworks. Extensive experiments across models and datasets demonstrate that our method further improves the efficiency of speculative decoding without altering the original inference framework, achieving an acceptance length of 5.92 and a 4.20$\times$ speedup on Qwen3-4B at a temperature of 1. Fig. \ref{fig:performance_comparision} presents the speedup comparison across different methods. Our contributions can be summarized as follows:
\begin{itemize}[leftmargin=*]
    \item We identify decoding policy as a learnable component in diffusion-based speculative decoding, showing that fixed block-size strategies are suboptimal across diverse inputs.
    \item We show that the optimal block size follows a structured local distribution, enabling efficient policy learning over a discrete action space.
    \item We propose BlockPilot, a lightweight instance-adaptive policy learning framework that predicts block size from the prefilling state, achieving consistent speedup gains with minimal overhead.
\end{itemize}

\section{Methodology}
\subsection{Problem Formulation}
Speculative decoding based on diffusion language models \cite{li2025diffuspecunlockingdiffusionlanguage,sandler2025specdiff2scalingdiffusiondrafter,samragh2025llmknowsfutureuncovering,chen2026dflash} is an efficient framework for accelerating autoregressive inference. It trains a lightweight diffusion language model as the draft model with a block size of $B$, and leverages a block-level diffusion mechanism to generate $B$ tokens in parallel. The target autoregressive model then verifies the generated sequence in parallel, alleviating the sequential bottleneck of autoregressive decoding. Formally, within a single speculative decoding step, the average per-token generation latency is defined as follows \cite{chen2026dflash}:
\begin{equation}
L(B) = \frac{T_{\text{draft}}(B) + T_{\text{verify}}(B)}{\tau(B)}
\end{equation}
where $L(B)$ denotes the average token generation latency under block size $B$, $T_{\text{draft}}(B)$ and $T_{\text{verify}}(B)$ represent the computational costs of the draft generation and verification stages, respectively, and $\tau(B)$ denotes the expected number of accepted tokens per verification step. Furthermore, the end-to-end speedup $\eta(B)$ is defined as:
\begin{equation}
\eta(B) = \frac{L_{\text{AR}}}{L(B)}
\end{equation}
where $L_{\text{AR}}$ denotes the average latency of standard autoregressive decoding. Since both draft generation and verification can be executed in a block-parallel manner, $T_{\text{draft}}(B)$ and $T_{\text{verify}}(B)$ typically increase sublinearly with $B$ and can often be approximated as near-constant overhead in practice \cite{hooper2025speed,stern2018blockwise,santilli2023accelerating}. This implies that $\tau(B)$ primarily governs efficiency, suggesting that an optimal inference block size $B^*$ maximizes $\tau(B)$ and improves end-to-end acceleration.

In common settings, inference typically uses the same block size $B$ as in training to maintain consistency between training and inference. However, this choice may be suboptimal at inference time, since the optimal block size can vary across input samples and deviate from the training configuration. To address this issue, we aim to adaptively determine a sample-specific optimal inference block size $B^*$ for each sample, instead of using a fixed training-time block size $B$. Specifically, $B^*$ is defined for each sample as the value that maximizes the expected acceptance length during speculative decoding, thereby improving inference efficiency.

\subsection{Key Findings}
To systematically analyze the impact of block size on speculative decoding performance, we perform an exhaustive sweep over candidate block sizes on multiple representative datasets. This allows us to directly compare decoding behavior under different levels of block-level parallelism. Specifically, we define the candidate set as $\mathcal{B} = \{1, 2, \ldots, 2B\}$, where $B$ denotes the block size used during training. For each input sample $x$, we define its optimal block size as follows:
\begin{equation}
\label{B*}
B^*(x) = \arg\max_{b \in \mathcal{B}} \tau(b; x)
\end{equation}
where $\tau(b; x)$ denotes the average number of accepted tokens for sample $x$ under block size $b$. The acceptance length determines the effective number of generated tokens per decoding step and therefore directly reflects how efficiently the draft tokens are utilized. It thus serves as a key metric for speculative decoding efficiency.

\begin{figure*}[t]
    \centering
    \begin{subfigure}[b]{0.32\textwidth}
        \centering
        \includegraphics[width=\textwidth]{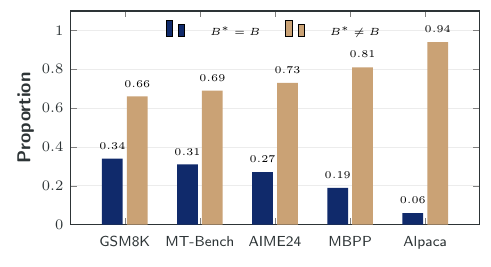}
        \caption{Proportion of samples with $B^*$ matching or mismatching $B$.}
        \label{fig:block_size_mismatch}
    \end{subfigure}
    \hfill
    \begin{subfigure}[b]{0.32\textwidth}
        \centering
        \includegraphics[width=\textwidth]{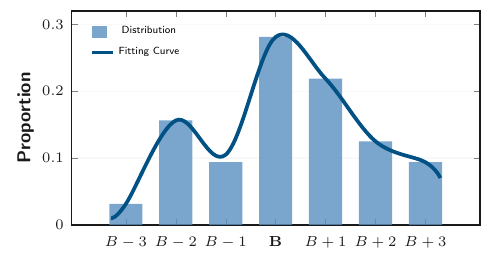}
        \caption{Unimodal distribution of $B^*$ peaking at the trained size $B$.}
        \label{fig:left_plot}
    \end{subfigure}
    \hfill
    \begin{subfigure}[b]{0.32\textwidth}
        \centering
        \includegraphics[width=\textwidth]{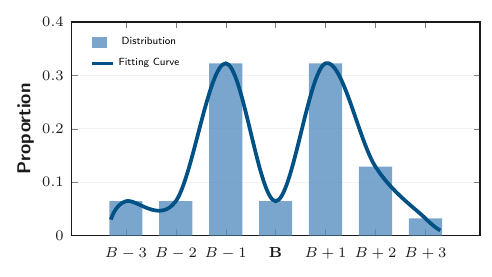}
        \caption{Symmetric bimodal distribution of $B^*$ near the trained size $B$.}
        \label{fig:right_plot}
    \end{subfigure}

    \caption{Analysis of optimal block size $B^*$. (a) Matching and mismatching proportions across datasets. (b-c) Distribution patterns demonstrating strong locality, where the range $[B-3, B+3]$ covers the optimal size for nearly all samples.}
    \label{fig:overall_label}
\end{figure*}

\paragraph{Finding I: Instance-wise Variability of Optimal Block Size.}
We first observe that the sample-wise optimal block size $B^*(x)$ varies significantly across samples, and does not necessarily match the fixed block size $B$ used during training. As shown in Fig.~\ref{fig:block_size_mismatch}, we report the proportions of samples satisfying $B^*(x)=B$ and $B^*(x)\neq B$ for each dataset. The results indicate that only a subset of samples align with the training configuration, while a substantial fraction prefer different block sizes at inference time, and this pattern is consistent across datasets.

These observations suggest that a fixed block size is insufficient to capture optimal decoding behavior across diverse inputs. The variation arises from differences in context structure and predictability: for some inputs, strong structural constraints from the prefilling stage enable high consistency over larger blocks, whereas for others, errors accumulate more rapidly as block size increases, resulting in shorter accepted lengths. This mismatch between the fixed training-time block size and sample-specific optimal decoding behavior motivates sample-adaptive block size selection at inference time.


\paragraph{Finding II: Local Interval Property of Optimal Block Size.}
Although the optimal block size varies across samples, its distribution exhibits a clear local interval structure. As shown in Fig.~\ref{fig:overall_label}, we observe two dominant patterns in the distribution, both indicating strong locality. Fig.~\ref{fig:left_plot} shows a unimodal distribution peaked at the training block size $B$, with probabilities rapidly decaying on both sides. Fig.~\ref{fig:right_plot} instead presents a bimodal but still localized pattern, with both peaks near $B$. Concretely, for almost all samples, the optimal block size satisfies as follows:
\begin{equation}
B^*(x) \in \{B-k, \ldots, B+k\}
\end{equation}
and it covers nearly all cases when $k=3$. Once the block size deviates from $B$ beyond this small range, the acceptance length drops sharply, making such choices rarely optimal. This indicates that the optimal block size exhibits strong locality across samples, with optimal solutions concentrated within a narrow region around the training block size.

This locality has important methodological implications. It reduces the unbounded search space over block sizes to a small discrete interval, within which the optimum almost always lies, significantly simplifying the problem. More importantly, it enables learning-based strategies that select block sizes from a limited candidate set rather than performing global search.

\paragraph{Finding III: Classification Formulation of Block Size Selection.}

Building on Finding II, which reveals a strong local interval structure of the optimal block size around the training block size, we formulate sample-adaptive block size selection as a structured classification task over a finite discrete space. Specifically, we restrict the candidate set to a local neighborhood $\{B-k, \ldots, B+k\}$ and learn a mapping function that predicts the optimal block size conditioned on the input sample. We use the predictive probability distribution of the last token after the prefilling stage as the input representation. Motivated by the causal structure of autoregressive decoding, the final token attends to the full context and thus provides a globally aggregated summary of the input sequence. Its predictive distribution captures contextual constraints, semantic consistency, and uncertainty in future generation, making it informative for block-size selection. We also explored using only the Top-$k$ probabilities as input, but this led to severe overfitting \cite{hinton2015distilling,berrada2018smooth}: the training accuracy reached around $80\%$, while the test accuracy remained only about $10\%$. Therefore, we retain the full predictive distribution to preserve richer information and improve generalization. Under this formulation, the probability distribution serves as an approximate sufficient statistic for optimal block size selection, enabling the classifier to make decisions with a single forward pass. Moreover, the proposed module can be seamlessly integrated into existing speculative decoding frameworks without modifying the generation pipeline.

\subsection{Training}
Following the classification formulation of the block size selection problem, the training objective is to learn a mapping function as follows:
\begin{equation}
f : p(x) \rightarrow B^*(x)
\end{equation}
where $p(x)$ denotes the predictive probability distribution of the last token after the prefilling stage for input $x$, and $B^*(x)$ denotes the corresponding optimal block size category. The training process consists of two components: supervised data construction and learning the block size predictor.

\paragraph{Supervised Data Construction.}
Since the optimal block size is not explicitly annotated, we construct supervised data via an offline enumeration strategy. For each input sample $x$, we first run the target model’s prefilling stage and extract the predictive distribution at the last position $p(x)$ as the input feature. Then, over the local candidate set $\mathcal{B} = \{B - k, \ldots, B + k\}$ determined by Finding II, we enumerate each candidate block size $b$ and execute full speculative decoding to measure the corresponding acceptance length $\tau(b; x)$. The block size that yields the maximum acceptance length is taken as the supervision signal, i.e., the optimal block size $B^*(x)$ defined in Eq.~\ref{B*}. Based on this procedure, we construct the training dataset $\mathcal{D}$ as follows:
\begin{equation}
\mathcal{D} = \{(p(x_i), B^*(x_i))\}_{i=1}^N
\end{equation}
In practice, the candidate set is small (typically $2k + 1 $), ensuring that the enumeration cost remains tractable. Since evaluations across different candidate block sizes are mutually independent, the process is highly parallelizable. This data construction procedure directly aligns with the optimization objective of speculative decoding, enabling the model to learn a mapping from decoding states to optimal decisions without relying on hand-crafted heuristics.

\paragraph{Block Size Predictor.}
We adopt a lightweight $n$-layer multilayer perceptron as a lightweight policy network. Since the input feature $p(x)$ is a high-level state representation compressed from the prefilling stage of the target model, sequence modeling architectures are unnecessary. Instead, we employ a simple discriminative model that is computationally efficient and easy to deploy. The model takes $p(x)$ as input and outputs logits over the candidate block size set $\mathcal{B}$. As $p(x)$ already encodes rich contextual information from the target model, a shallow architecture is sufficient to achieve strong performance while introducing only limited latency. The output distribution is defined via a softmax over candidate block sizes:
\begin{equation}
P(b \mid x) = \frac{\exp(o_b)}{\sum_{b' \in \mathcal{B}} \exp(o_{b'})}
\end{equation}
where $o_b$ denotes the logit corresponding to the candidate block size $b \in \mathcal{B}$. The model is trained by minimizing the standard cross-entropy loss over the training dataset:
\begin{equation}
L = -\frac{1}{N} \sum_{i=1}^{N} \log P\!\left(B^*(x_i)\mid x_i\right)
\end{equation}
where $N$ denotes the number of training samples and $B^*(x_i)$ represents the ground-truth optimal block size for sample $x_i$. This objective encourages the model to assign higher probability mass to the optimal block size, thereby learning a direct mapping from decoding states to optimal decisions. Owing to the compact design of the predictor, both training and inference incur relatively low computational overhead.

\begin{figure}[t]
    \centering
    \includegraphics[width=1\textwidth]{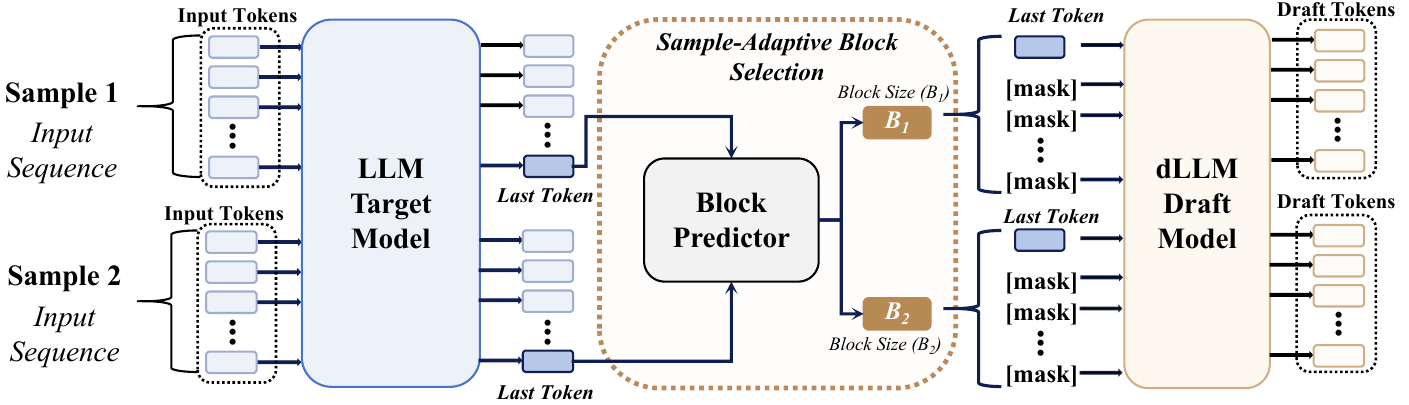}
    \caption{Overview of the BlockPilot inference pipeline. Given an input sequence, the target LLM performs prefilling and produces the predictive distribution of the last token, which serves as a compact representation of the decoding state. This distribution is then fed into a lightweight block size predictor to determine an instance-specific block size. Based on the predicted block size, the diffusion-based draft model generates a block of draft tokens in parallel.}
    \label{fig:inference}
\end{figure}

\subsection{Inference}

\begin{wraptable}{r}{0.46\textwidth}
\vspace{-8pt}
\centering
\small
\setlength{\tabcolsep}{3.2pt} 
\renewcommand{\arraystretch}{1.15} 
\caption{Overhead analysis of the block size predictor compared to backbone models.}
\label{tab:model_efficiency}
\resizebox{0.46\textwidth}{!}{%
\begin{tabular}{l c rr r}
\toprule
\multirow{2}{*}{\textbf{Model}} & \multirow{2}{*}{\textbf{\#Params}} & \multicolumn{2}{c}{\textbf{Memory}} & \multirow{2}{*}{\textbf{Latency}} \\
\cmidrule(l{2pt}r{2pt}){3-4} 
& & \textbf{Act.} & \textbf{Param.} & \\
\midrule
Qwen3-4B              & 4.41B & 6.53 GB & 8.62 GB  & 183.48 ms \\
\rowcolor{gray!10}
Qwen3-8B              & 8.19B & 8.16 GB & 16.00 GB & 278.37 ms \\
Llama-3.1-8B-Instruct & 8.03B & 7.10 GB & 15.68 GB & 272.14 ms \\
\midrule
\rowcolor[HTML]{ecf2f9}
\textit{Block Size Predictor} & 0.32B & 0.02 MB & 0.62 GB & 7.34 ms \\
\bottomrule
\end{tabular}%
}
\vspace{-8pt}
\end{wraptable}
During inference, we integrate the learned block size predictor into the speculative decoding pipeline to enable sample-wise adaptive block size selection. Given an input sample $x$, we run the prefilling stage of the target model and extract the predictive distribution at the last position, denoted as $p(x)$. The predictor then produces logits over the candidate set $\mathcal{B}$, and the block size is determined as follows:
\begin{equation}
\hat{B}(x) = \arg\max_{b \in \mathcal{B}} f(p(x))_b
\end{equation}
where $f(p(x))_b$ denotes the predicted score for candidate block size $b$. Once $\hat{B}(x)$ is obtained, it is fixed for the entire subsequent speculative decoding process, including both draft generation and target model verification. Fig. \ref{fig:inference} presents an overview of the inference process. Compared to fixed block size strategies, this approach adaptively selects a more suitable level of parallel generation conditioned on each input sample. Notably, the block size prediction is performed only once after the prefilling stage, and the predictor itself is implemented as a lightweight network, resulting in minimal computational overhead. Table~\ref{tab:model_efficiency} compares the predictor with the backbone model in terms of parameter size, memory footprint, and inference latency, showing that the predictor introduces only millisecond-level latency. Although the predictor introduces a small additional memory footprint, this cost is minor compared with the backbone models and is well compensated by the resulting speedup. Therefore, the small additional memory overhead represents a favorable trade-off for improving decoding efficiency.

\begin{table*}[t]
\centering
\scriptsize
\setlength{\tabcolsep}{3.0pt}
\renewcommand{\arraystretch}{1.14}
\caption{Speedup ratios and average acceptance length $\tau$ on Qwen3 models across Math, Code, and Chat benchmarks. Q3-4B and Q3-8B denote Qwen3-4B and Qwen3-8B, respectively. DFlash$(n)$ denotes DFlash with block size $n$.}
\label{tab:qwen3_speedup_acceptance}
\resizebox{\textwidth}{!}{%
\begin{tabular}{@{}ll*{8}{cc}@{}}
\toprule
\multirow{2}{*}{\textbf{Model}}
& \multirow{2}{*}{\textbf{Method}}
& \multicolumn{6}{c}{\textbf{Math}}
& \multicolumn{6}{c}{\textbf{Code}}
& \multicolumn{2}{c}{\textbf{Chat}}
& \multicolumn{2}{c}{\textbf{Overall}} \\
\cmidrule(lr){3-8}
\cmidrule(lr){9-14}
\cmidrule(lr){15-16}
\cmidrule(l){17-18}
&
& \multicolumn{2}{c}{GSM8K}
& \multicolumn{2}{c}{MATH-500}
& \multicolumn{2}{c}{AIME24}
& \multicolumn{2}{c}{HumanEval}
& \multicolumn{2}{c}{MBPP}
& \multicolumn{2}{c}{SWE-Bench}
& \multicolumn{2}{c}{MT-Bench}
& \multicolumn{2}{c}{Avg.} \\
\midrule
\multicolumn{2}{@{}l}{\textit{Temperature} $=0$}
& \textit{Speedup} & $\tau$
& \textit{Speedup} & $\tau$
& \textit{Speedup} & $\tau$
& \textit{Speedup} & $\tau$
& \textit{Speedup} & $\tau$
& \textit{Speedup} & $\tau$
& \textit{Speedup} & $\tau$
& \textit{Speedup} & $\tau$ \\
\midrule
\multirow{6}{*}{Q3-4B}
& EAGLE-3     & 1.99$\times$ & 3.30 & 1.83$\times$ & 3.08 & 1.79$\times$ & 3.05 & 1.84$\times$ & 3.05 & 1.78$\times$ & 2.95 & 1.39$\times$ & 2.48 & 1.74$\times$ & 3.02 & 1.70$\times$ & 2.95 \\
& DFlash (4)  & 2.29$\times$ & 3.32 & 2.20$\times$ & 3.48 & 2.12$\times$ & 3.45 & 2.12$\times$ & 3.22 & 2.16$\times$ & 3.17 & 1.59$\times$ & 2.58 & 1.45$\times$ & 2.85 & 1.99$\times$ & 3.15 \\
& DFlash (8)  & 3.56$\times$ & 5.16 & 3.59$\times$ & 5.68 & 3.47$\times$ & 5.64 & 3.19$\times$ & 4.84 & 3.22$\times$ & 4.71 & 1.95$\times$ & 3.17 & 1.99$\times$ & 3.91 & 2.99$\times$ & 4.73 \\
& DFlash (16) & 4.55$\times$ & 6.59 & 5.25$\times$ & 8.31 & 4.58$\times$ & 7.45 & 4.28$\times$ & 6.51 & 4.26$\times$ & 6.24 & 2.33$\times$ & 3.78 & 2.69$\times$ & 5.29 & 3.99$\times$ & 6.31 \\
& DFlash (32) & 2.92$\times$ & 4.24 & 3.22$\times$ & 5.09 & 2.99$\times$ & 4.86 & 2.56$\times$ & 3.89 & 2.71$\times$ & 3.97 & 1.69$\times$ & 2.74 & 1.62$\times$ & 3.19 & 2.53$\times$ & 4.00 \\
& \textbf{BlockPilot} & \textbf{4.76$\times$} & \textbf{6.90} & \textbf{5.45$\times$} & \textbf{8.62} & \textbf{4.80$\times$} & \textbf{7.82} & \textbf{4.46$\times$} & \textbf{6.77} & \textbf{4.38$\times$} & \textbf{6.42} & \textbf{2.45$\times$} & \textbf{3.98} & \textbf{2.85$\times$} & \textbf{5.62} & \textbf{4.17$\times$} & \textbf{6.59} \\
\midrule
\multirow{6}{*}{Q3-8B}
& EAGLE-3     & 1.94$\times$ & 3.23 & 1.81$\times$ & 3.02 & 1.79$\times$ & 3.00 & 1.89$\times$ & 3.17 & 1.69$\times$ & 2.82 & 1.47$\times$ & 2.45 & 1.63$\times$ & 2.83 & 1.75$\times$ & 2.93 \\
& DFlash (4)  & 2.45$\times$ & 3.29 & 2.50$\times$ & 3.47 & 2.55$\times$ & 3.44 & 2.45$\times$ & 3.33 & 2.32$\times$ & 3.21 & 1.96$\times$ & 2.58 & 1.51$\times$ & 2.58 & 2.25$\times$ & 3.13 \\
& DFlash (8)  & 3.76$\times$ & 5.04 & 4.09$\times$ & 5.67 & 4.22$\times$ & 5.69 & 3.82$\times$ & 5.19 & 3.51$\times$ & 4.86 & 2.45$\times$ & 3.23 & 2.01$\times$ & 3.45 & 3.41$\times$ & 4.73 \\
& DFlash (16) & 4.89$\times$ & 6.56 & 5.98$\times$ & 8.29 & 5.57$\times$ & 7.51 & 4.98$\times$ & 6.76 & 4.33$\times$ & 5.99 & 2.72$\times$ & 3.58 & 2.48$\times$ & 4.25 & 4.42$\times$ & 6.13 \\
& DFlash (32) & 4.47$\times$ & 5.99 & 5.66$\times$ & 7.84 & 5.20$\times$ & 7.01 & 4.66$\times$ & 6.32 & 3.92$\times$ & 5.42 & 2.50$\times$ & 3.29 & 2.37$\times$ & 4.06 & 4.11$\times$ & 5.70 \\
& \textbf{BlockPilot} & \textbf{5.02$\times$} & \textbf{6.73} & \textbf{6.17$\times$} & \textbf{8.56} & \textbf{6.16$\times$} & \textbf{8.30} & \textbf{5.21$\times$} & \textbf{7.07} & \textbf{4.53$\times$} & \textbf{6.26} & \textbf{2.96$\times$} & \textbf{3.90} & \textbf{2.55$\times$} & \textbf{4.37} & \textbf{4.66$\times$} & \textbf{6.46} \\
\midrule
\multicolumn{2}{@{}l}{\textit{Temperature} $=1$}
& \textit{Speedup} & $\tau$
& \textit{Speedup} & $\tau$
& \textit{Speedup} & $\tau$
& \textit{Speedup} & $\tau$
& \textit{Speedup} & $\tau$
& \textit{Speedup} & $\tau$
& \textit{Speedup} & $\tau$
& \textit{Speedup} & $\tau$ \\
\midrule
\multirow{6}{*}{Q3-4B}
& EAGLE-3     & 1.89$\times$ & 3.22 & 1.75$\times$ & 2.99 & 1.64$\times$ & 2.79 & 1.74$\times$ & 3.01 & 1.69$\times$ & 2.89 & 1.47$\times$ & 2.31 & 1.77$\times$ & 2.99 & 1.70$\times$ & 2.88 \\
& DFlash (4)  & 2.32$\times$ & 3.17 & 2.34$\times$ & 3.26 & 2.08$\times$ & 2.90 & 2.29$\times$ & 3.11 & 2.25$\times$ & 3.10 & 1.79$\times$ & 2.36 & 1.62$\times$ & 2.74 & 2.10$\times$ & 2.95 \\
& DFlash (8)  & 3.60$\times$ & 4.91 & 3.71$\times$ & 5.16 & 2.97$\times$ & 4.13 & 3.37$\times$ & 4.58 & 3.21$\times$ & 4.42 & 2.17$\times$ & 2.86 & 2.24$\times$ & 3.79 & 3.04$\times$ & 4.26 \\
& DFlash (16) & 4.36$\times$ & 5.95 & 4.83$\times$ & 6.72 & 3.66$\times$ & 5.10 & 4.32$\times$ & 5.87 & 4.02$\times$ & 5.54 & 2.33$\times$ & 3.07 & 3.10$\times$ & 5.23 & 3.80$\times$ & 5.35 \\
& DFlash (32) & 2.83$\times$ & 3.87 & 3.35$\times$ & 4.66 & 2.79$\times$ & 3.88 & 2.76$\times$ & 3.76 & 2.69$\times$ & 3.71 & 1.91$\times$ & 2.52 & 1.91$\times$ & 3.22 & 2.61$\times$ & 3.66 \\
& \textbf{BlockPilot} & \textbf{4.77$\times$} & \textbf{6.51} & \textbf{5.41$\times$} & \textbf{7.53} & \textbf{4.02$\times$} & \textbf{5.60} & \textbf{4.74$\times$} & \textbf{6.45} & \textbf{4.59$\times$} & \textbf{6.33} & \textbf{2.58$\times$} & \textbf{3.39} & \textbf{3.31$\times$} & \textbf{5.59} & \textbf{4.20$\times$} & \textbf{5.92} \\
\midrule
\multirow{6}{*}{Q3-8B}
& EAGLE-3     & 1.87$\times$ & 3.12 & 1.73$\times$ & 2.91 & 1.63$\times$ & 2.74 & 1.75$\times$ & 3.05 & 1.64$\times$ & 2.74 & 1.33$\times$ & 2.11 & 1.58$\times$ & 2.70 & 1.65$\times$ & 2.77 \\
& DFlash (4)  & 2.34$\times$ & 3.17 & 2.25$\times$ & 3.22 & 2.05$\times$ & 2.92 & 2.25$\times$ & 3.05 & 2.23$\times$ & 3.03 & 1.66$\times$ & 2.20 & 1.44$\times$ & 2.45 & 2.03$\times$ & 2.86 \\
& DFlash (8)  & 3.48$\times$ & 4.72 & 3.53$\times$ & 5.06 & 2.92$\times$ & 4.17 & 3.26$\times$ & 4.41 & 3.20$\times$ & 4.36 & 1.94$\times$ & 2.57 & 1.93$\times$ & 3.28 & 2.89$\times$ & 4.08 \\
& DFlash (16) & 4.28$\times$ & 5.81 & 4.62$\times$ & 6.62 & 3.49$\times$ & 4.98 & 4.19$\times$ & 5.68 & 3.98$\times$ & 5.41 & 2.09$\times$ & 2.77 & 2.20$\times$ & 3.75 & 3.55$\times$ & 5.00 \\
& DFlash (32) & 4.10$\times$ & 5.57 & 4.41$\times$ & 6.32 & 3.39$\times$ & 4.84 & 3.77$\times$ & 5.11 & 3.68$\times$ & 5.00 & 2.01$\times$ & 2.67 & 2.23$\times$ & 3.79 & 3.37$\times$ & 4.75 \\
& \textbf{BlockPilot} & \textbf{4.69$\times$} & \textbf{6.37} & \textbf{5.06$\times$} & \textbf{7.25} & \textbf{3.84$\times$} & \textbf{5.47} & \textbf{4.65$\times$} & \textbf{6.30} & \textbf{4.54$\times$} & \textbf{6.17} & \textbf{2.30$\times$} & \textbf{3.05} & \textbf{2.48$\times$} & \textbf{4.23} & \textbf{3.94$\times$} & \textbf{5.55} \\
\bottomrule
\end{tabular}%
}
\end{table*}

\section{Experiments}
\label{Experiments}
\subsection{Experimental Setup}
\paragraph{Models and Benchmarks.}
We evaluate our method on four representative LLMs: Qwen3-4B, Qwen3-8B \cite{yang2025qwen3}, Llama-3.1-8B-Instruct \cite{grattafiori2024llama}, and Qwen3-Coder-30B-A3B \cite{cao2026qwen3}, spanning diverse scales and domains, including both general-purpose instruction-tuned and code-specialized models. We consider benchmarks across three task categories. For mathematical reasoning, we use GSM8K~\citep{cobbe2021training}, MATH-500~\citep{lightman2023let}, and AIME24~\citep{AIME}. For code generation and software engineering, we adopt HumanEval~\citep{chen2021evaluating}, MBPP~\citep{austin2021program}, and SWE-Bench~\citep{jimenez2023swe}. For conversational generation, we use MT-Bench~\citep{zheng2023judging}. Together, these benchmarks cover Math, Code, and Chat scenarios.

\paragraph{Baselines.}
We compare our method with standard autoregressive decoding (baseline), the most classical speculative decoding method with an autoregressive drafter, EAGLE-3~\citep{li2025eagle3scalinginferenceacceleration}, and the state-of-the-art (SOTA) diffusion-based counterpart, DFlash~\citep{chen2026dflash}. Here, DFlash$(n)$ denotes DFlash with block size $n$.

\paragraph{Metrics.}
Since our method preserves the exact output distribution of the target model under speculative decoding, generation quality remains unchanged. Therefore, we focus on efficiency metrics, measured using the following metrics:
\begin{itemize}[leftmargin=*]
    \item \textbf{Average Acceptance Length $\tau$:} The average number of tokens accepted from the draft model per drafting-verification cycle.
    \item \textbf{Speedup Ratio:} The ratio of inference time for standard autoregressive decoding to that for different speculative decoding methods.
\end{itemize}

\paragraph{Implementation Details.}
Our experiments are conducted using the PyTorch framework \citep{paszke2019pytorch} and the Hugging Face Transformers library \citep{wolf2020transformers}, running on NVIDIA H100 80GB GPUs. We construct the training dataset from ShareGPT \cite{chen2024sharegpt4v}, WSC \cite{levesque2012winograd}, and COPA \cite{roemmele2011choice}. We train the model for 100 epochs using the Adam optimizer with a learning rate of $1e^{-5}$. The predictor network consists of two layers with a hidden dimension of 2048, and the default value of the hyperparameter $k$ is set to 2.

\subsection{Main Results}

Table~\ref{tab:qwen3_speedup_acceptance} presents the main results on the Qwen3 series models under both deterministic decoding and stochastic sampling settings. Notably, these improvements are achieved without modifying either the draft or target models and introduce only negligible additional latency.  Overall, BlockPilot consistently achieves the best performance across models, temperatures, and benchmark categories. On Qwen3-4B, it reaches average speedups of 4.17$\times$ and 4.20$\times$ under temperature $=0$ and temperature $=1$, respectively. On Qwen3-8B, it achieves 4.66$\times$ and 3.94$\times$ under the two settings. These results outperform EAGLE-3 and all fixed-block DFlash variants, demonstrating the effectiveness of sample-adaptive block size selection. This indicates that the proposed adaptive strategy generalizes well across different model scales and decoding regimes.

Compared with the strongest fixed-block DFlash baseline, BlockPilot further improves both speedup and average acceptance length $\tau$. Under temperature $=0$, DFlash(16) is generally the best fixed-block baseline. On Qwen3-4B, our method improves the average speedup from 3.99$\times$ to 4.17$\times$ and increases $\tau$ from 6.31 to 6.59. On Qwen3-8B, the speedup rises from 4.42$\times$ to 4.66$\times$, while $\tau$ improves from 6.13 to 6.46. Similar gains are observed under temperature $=1$, where our method improves the average speedup from 3.80$\times$ to 4.20$\times$ on Qwen3-4B and from 3.55$\times$ to 3.94$\times$ on Qwen3-8B. These results reveal the limitation of fixed-block inference: larger blocks provide more parallelism, but do not always yield better acceleration. For example, DFlash(32) often underperforms DFlash(16), suggesting that overly large blocks may reduce acceptance due to accumulated drafting errors. In contrast, our method selects a suitable block size for each input sample, better balancing drafting parallelism and verification acceptance.

Across Math, Code, and Chat benchmarks, our method remains consistently effective. The gains are also preserved under temperature $=1$, where generation is more stochastic and draft-token acceptance becomes more challenging. This suggests that the last-token predictive distribution after prefilling provides a useful signal for adaptive block size selection. More results are provided in Appendix~\ref{app:main}.



\subsection{Ablation Studies}
We conduct ablation studies on Qwen3-4B under a zero-temperature setting to analyze the key design choices of our method, including the predictor architecture, input preprocessing, and candidate interval radius, in order to better understand their impact on performance.

\paragraph{Predictor Configuration.}

\begin{wraptable}{r}{0.46\textwidth}
\vspace{-8pt}
\centering
\scriptsize
\setlength{\tabcolsep}{3.2pt}
\renewcommand{\arraystretch}{1.03}
\caption{Speedup and acceptance length ($\tau$) under different predictor configurations.}
\label{tab:predictor_config}
\resizebox{0.46\textwidth}{!}{%
\begin{tabular}{lcccccc}
\toprule
\multirow{2}{*}{Setting}
& \multicolumn{2}{c}{GSM8K}
& \multicolumn{2}{c}{HumanEval}
& \multicolumn{2}{c}{MT-Bench} \\
\cmidrule(lr){2-3}\cmidrule(lr){4-5}\cmidrule(lr){6-7}
& \textit{Speedup} & $\tau$
& \textit{Speedup} & $\tau$
& \textit{Speedup} & $\tau$ \\
\midrule
\multicolumn{7}{c}{\textit{Hidden width} ($L=2$)} \\
\midrule
$D=1024$ & 4.73$\times$ & 6.86 & 4.41$\times$ & 6.70 & 2.85$\times$ & 5.62 \\
$D=2048$ & 4.76$\times$ & 6.90 & 4.46$\times$ & 6.77 & 2.85$\times$ & 5.62 \\
$D=4096$ & 4.76$\times$ & 6.90 & 4.46$\times$ & 6.77 & 2.85$\times$ & 5.62 \\
\midrule
\multicolumn{7}{c}{\textit{Predictor depth} ($D=2048$)} \\
\midrule
$L=1$ & 4.71$\times$ & 6.83 & 4.42$\times$ & 6.72 & 2.84$\times$ & 5.59 \\
$L=2$ & 4.76$\times$ & 6.90 & 4.46$\times$ & 6.77 & 2.85$\times$ & 5.62 \\
$L=3$ & 4.65$\times$ & 6.74 & 4.67$\times$ & 7.10 & 2.85$\times$ & 5.62 \\
\bottomrule
\end{tabular}%
}
\vspace{-8pt}
\end{wraptable}

To further understand the impact of predictor design, we report results under different configurations in Table~\ref{tab:predictor_config}. Fixing the depth to $L=2$, increasing the hidden size from $D=1024$ to $D=2048$ improves both speedup and acceptance length $\tau$, while further scaling to $D=4096$ yields negligible gains. This suggests that a moderate hidden width is sufficient to capture the mapping from prefilling distributions to block size decisions. We then fix the hidden size to $D=2048$ and vary the depth. Compared to a single-layer predictor, a two-layer model achieves better overall performance, indicating the benefit of moderate nonlinearity. Increasing the depth to $L=3$ leads to gains on HumanEval but slight drops on GSM8K, yielding no clear overall advantage. Therefore, we adopt a two-layer MLP with hidden size 2048 as the default predictor, striking a good balance between efficiency and performance.

\paragraph{Candidate Interval Radius.}

\begin{wraptable}{r}{0.48\textwidth}
\vspace{-10pt}
\centering
\scriptsize
\setlength{\tabcolsep}{4.0pt}
\renewcommand{\arraystretch}{1.05}
\caption{Effect of the candidate interval radius $k$ on speedup and acceptance length ($\tau$).}
\label{tab:k_analysis}
\resizebox{0.48\textwidth}{!}{%
\begin{tabular}{lcccccc}
\toprule
\multirow{2}{*}{Setting}
& \multicolumn{2}{c}{GSM8K}
& \multicolumn{2}{c}{HumanEval}
& \multicolumn{2}{c}{MT-Bench} \\
\cmidrule(lr){2-3}\cmidrule(lr){4-5}\cmidrule(lr){6-7}
& \textit{Speedup} & $\tau$
& \textit{Speedup} & $\tau$
& \textit{Speedup} & $\tau$ \\
\midrule
$k=1$ & 4.67$\times$ & 6.77 & 4.44$\times$ & 6.75 & 2.78$\times$ & 5.47 \\
$k=2$ & 4.76$\times$ & 6.90 & 4.46$\times$ & 6.77 & 2.85$\times$ & 5.62 \\
$k=3$ & 4.64$\times$ & 6.73 & 4.39$\times$ & 6.67 & 2.87$\times$ & 5.66 \\
\bottomrule
\end{tabular}%
}
\vspace{-10pt}
\end{wraptable}


We additionally study the effect of the candidate interval radius $k$ in Table~\ref{tab:k_analysis}, where $k$ defines the local block-size search range ${B-k,\ldots,B+k}$ centered at the training block size $B$. This hyperparameter balances candidate coverage and prediction difficulty. A smaller radius ($k=1$) restricts the search space, simplifying prediction but potentially missing better block-size choices. Increasing the radius to $k=2$ improves both speedup and acceptance length on GSM8K and HumanEval, and also brings gains on MT-Bench, suggesting that a moderately wider interval provides more useful candidates during inference. When further expanded to $k=3$, performance does not improve on most metrics, likely because the larger candidate set introduces more competing choices and increases prediction difficulty. Therefore, we set $k=2$ as the default configuration, which achieves a favorable balance between candidate coverage and prediction difficulty.


\paragraph{Predictor Input Preprocessing.}

\begin{wraptable}{r}{0.48\textwidth}
\vspace{-10pt}
\centering
\scriptsize
\setlength{\tabcolsep}{4.0pt}
\renewcommand{\arraystretch}{1.05}
\caption{Effect of predictor input preprocessing on speedup and acceptance length ($\tau$).}
\label{tab:input_preprocessing}
\resizebox{0.48\textwidth}{!}{%
\begin{tabular}{lcccccc}
\toprule
\multirow{2}{*}{Setting}
& \multicolumn{2}{c}{GSM8K}
& \multicolumn{2}{c}{HumanEval}
& \multicolumn{2}{c}{MT-Bench} \\
\cmidrule(lr){2-3}\cmidrule(lr){4-5}\cmidrule(lr){6-7}
& \textit{Speedup} & $\tau$
& \textit{Speedup} & $\tau$
& \textit{Speedup} & $\tau$ \\
\midrule
Softmax        & 4.55$\times$ & 6.59 & 4.28$\times$ & 6.51 & 2.69$\times$ & 5.29 \\
Normalization  & 4.64$\times$ & 6.73 & 4.44$\times$ & 6.75 & 2.83$\times$ & 5.58 \\
\textbf{Ours}  & \textbf{4.76$\times$} & \textbf{6.90} & \textbf{4.46$\times$} & \textbf{6.77} & \textbf{2.85$\times$} & \textbf{5.62} \\
\bottomrule
\end{tabular}%
}
\vspace{-10pt}
\end{wraptable}

We further study how preprocessing the predictor input affects performance in Table~\ref{tab:input_preprocessing}. The predictor input is the predictive distribution of the last token after prefilling. Compared with directly using this distribution, both normalization and softmax preprocessing consistently reduce speedup and acceptance length $\tau$. This suggests that the raw prefilling distribution already encodes sufficient confidence information for block-size prediction, while additional preprocessing may alter or partially suppress this signal. Therefore, we use the original prefilling distribution as the predictor input.

\section{Related Work}
\paragraph{Speculative Decoding.}
Speculative decoding improves the efficiency of large language model inference by alleviating the inherent sequential constraint of autoregressive generation. Early approaches~\cite{leviathan2023fastinferencetransformersspeculative} introduce a lightweight draft model to generate candidate token sequences, which are then verified in parallel by a larger target model. Building on this idea, Medusa~\cite{cai2024medusasimplellminference} eliminates the need for an external draft model by equipping the base LLM with multiple prediction heads, enabling parallel candidate generation via a tree-attention mechanism. More recent work in the EAGLE family~\cite{li2025eaglespeculativesamplingrequires,li2024eagle2fasterinferencelanguage,li2025eagle3scalinginferenceacceleration} further explores feature-level speculative decoding by leveraging intermediate hidden states of the target model. EAGLE-1 predicts future hidden-state distributions to improve token acceptance rates, while EAGLE-2 introduces adaptive drafting structures to better balance efficiency and accuracy. EAGLE-3 further improves training scalability and generalization across model sizes.

\paragraph{Diffusion Language Models.}
Diffusion-based large language models (dLLMs) offer a non-autoregressive paradigm via parallel masked token prediction. LLaDA~\cite{nie2025largelanguagediffusionmodels} first scales dLLMs to the billion-parameter level, achieving performance comparable to LLaMA-3.1-8B~\cite{grattafiori2024llama}. However, fully parallel diffusion models are limited by fixed-length generation and inefficient KV cache usage. Block diffusion models~\cite{arriola2025blockdiffusioninterpolatingautoregressive} address this by denoising sequences in blocks, combining parallelism with autoregressive structure. Building on this, Fast-dLLM v2~\cite{wu2025fastdllmv2efficientblockdiffusion} and SDAR~\cite{cheng2025sdarsynergisticdiffusionautoregressionparadigm} convert pretrained autoregressive LLMs into block-diffusion variants, enabling parallel generation with competitive quality on specific tasks. Nevertheless, dLLMs still lag behind state-of-the-art autoregressive models and require many denoising steps, limiting inference efficiency.

\paragraph{Diffusion-based Speculative Decoding.}
Recent work explores diffusion-based draft generation for speculative decoding. TiDAR~\cite{liu2025tidarthinkdiffusiontalk} combines diffusion and autoregressive objectives for parallel drafting, but still fails to achieve lossless generation quality. Another line of work adapts autoregressive models for diffusion-style drafting. \cite{samragh2025llmknowsfutureuncovering} use a LoRA adapter to enable parallel drafting from implicit future-token signals, though with limited effectiveness. DiffuSpec~\cite{li2025diffuspecunlockingdiffusionlanguage} and SpecDiff-2~\cite{sandler2025specdiff2scalingdiffusiondrafter} rely on large pre-trained diffusion LMs with search or alignment to improve acceptance, but require 7B-scale draft models, incurring substantial memory and latency overhead that hinders deployment. All the above methods remain difficult to apply efficiently in practice. Recently, DFlash~\cite{chen2026dflash} achieves a practical state-of-the-art (SOTA) method for block diffusion-based speculative decoding by injecting target-model hidden states into the diffusion drafter, substantially improving draft quality, acceptance length, and inference speed.

\section{Conclusion}
In this paper, we revisit diffusion-based speculative decoding and identify block size as a key factor in inference efficiency. We show that the optimal block size varies across samples but exhibits strong locality around the training configuration, reducing it to a small structured decision problem. Based on this, we propose BlockPilot, a  sample-adaptive predictor that uses the prefilling-stage predictive distribution to select block size from a local candidate set. The method is applied once per sample and integrates seamlessly into existing frameworks. Our method demonstrates that the decoding policy, rather than the model architecture alone, plays a critical role in inference efficiency.

\bibliographystyle{plainnat}
\bibliography{neurips_2025}

\appendix
\newpage

\section{Theoretical Analysis of Sample-Adaptive Block Size Selection}
\label{Theoretical Analysis}

\subsection{Acceptance Length as a Prefix-Survival Process}

\paragraph{Prefix-survival identity.}
We first formalize the expected acceptance length in speculative decoding from a prefix-survival perspective. Given an input sequence $x$ and an inference block size $b$, let $L_b(x)$ denote the number of consecutive draft tokens accepted by the target model in one speculative decoding step. The expected acceptance length is defined as
\begin{equation}
\tau(b;x)
=
\mathbb{E}[L_b(x)].
\end{equation}

Since $L_b(x)$ is a non-negative integer-valued random variable bounded by the block size $b$, its expectation can be decomposed into the sum of survival probabilities:
\begin{equation}
\tau(b;x)
=
\sum_{i=1}^{b}
\mathbb{P}(L_b(x)\ge i).
\end{equation}
This identity provides a prefix-level view of speculative decoding: each term measures the probability that the verified prefix survives up to at least position $i$.

In speculative decoding, the target model accepts only the longest consistent prefix of the drafted block. Therefore, accepting at least $i$ draft tokens requires that the first $i$ drafted tokens are all accepted. Define the prefix-consistency event
\begin{equation}
A_i
=
\{\text{the first $i$ drafted tokens are all accepted}\}.
\end{equation}
Then
\begin{equation}
\mathbb{P}(L_b(x)\ge i)
=
\mathbb{P}(A_i).
\end{equation}

We define the conditional acceptance probability at position $j$ as
\begin{equation}
q_j(x,b)
=
\mathbb{P}
\left(
\text{the $j$-th drafted token is accepted}
\mid
L_b(x)\ge j-1, x, b
\right).
\end{equation}
By the chain rule of probability, the probability that the first $i$ drafted tokens all survive verification is
\begin{equation}
\mathbb{P}(A_i)
=
\prod_{j=1}^{i}
q_j(x,b).
\end{equation}
Substituting this expression into the survival identity yields
\begin{equation}
\label{eq:tau_prefix_survival}
\tau(b;x)
=
\sum_{i=1}^{b}
\prod_{j=1}^{i}
q_j(x,b).
\end{equation}

Equation~\eqref{eq:tau_prefix_survival} shows that the acceptance length can be interpreted as the expected stopping time of a truncated prefix-survival process. This formulation is more informative than treating acceptance length as a black-box empirical statistic: the block size determines the truncation horizon, while the actual contribution of each additional drafted position is governed by the corresponding prefix survival probability.

\paragraph{Block-size-dependent acceptance.}
The conditional probability $q_j(x,b)$ should not be viewed as an intrinsic constant independent of the inference strategy. In block-level diffusion drafting, the draft model generates a set of mutually dependent tokens within a single block. Increasing the block size enlarges the maximum possible accepted length, but it may also make each prefix harder to verify because the proposal distribution must remain coherent over a longer span.

Therefore, increasing $b$ induces two competing effects:
\begin{equation}
\tau(b;x)
=
\sum_{i=1}^{b}
\underbrace{
\prod_{j=1}^{i}
q_j(x,b)
}_{\text{prefix survival probability}}.
\end{equation}
A larger block size increases the summation horizon by adding more candidate positions. However, the same change may alter the conditional acceptance probabilities inside each multiplicative term. Since prefix survival is multiplicative, even a mild degradation in $q_j(x,b)$ can be amplified over longer prefixes. Hence, the expected acceptance length is not determined solely by the number of drafted tokens, but by the interaction between block length and prefix survival.

\subsection{Locality Induced by Predictability and Block-Size Retention}

\paragraph{Predictability--retention decomposition.}
Let $B$ denote the block size used to train the diffusion draft model. Since the draft model is optimized to generate coherent token blocks under this configuration, its proposal distribution is naturally best calibrated near $B$. When the inference block size $b$ deviates substantially from $B$, the draft distribution may become less aligned with the target model, which can reduce proposal quality and lower acceptance probabilities.

To capture this mechanism in a simple analytical form, we decompose the conditional acceptance probability as
\begin{equation}
q_j(x,b)
=
\gamma_j(x)\, r(b;B),
\end{equation}
where $\gamma_j(x)\in(0,1]$ represents the intrinsic predictability of sample $x$ at position $j$, and $r(b;B)\in(0,1]$ measures the retention of proposal quality under inference block size $b$ relative to the training block size $B$. This decomposition separates two effects: sample-dependent predictability and block-size-induced proposal degradation.

A convenient retention model is
\begin{equation}
r(b;B)
=
\exp\{-\alpha(b-B)^2\},
\quad
\alpha>0.
\end{equation}
This function reaches its maximum at the training block size $B$ and gradually penalizes deviations from it. We emphasize that this retention function is used only as an analytical model to explain the observed locality of optimal block sizes, rather than as an additional component required by the algorithm.

Substituting the decomposition into Eq.~\eqref{eq:tau_prefix_survival}, we obtain
\begin{equation}
\label{eq:tau_predictability_retention}
\tau(b;x)
=
\sum_{i=1}^{b}
r(b;B)^i
\prod_{j=1}^{i}
\gamma_j(x).
\end{equation}
This expression makes the local structure explicit. Although increasing $b$ adds more candidate positions, deviations from $B$ reduce the prefix survival terms through the factor $r(b;B)^i$. The effect becomes stronger for longer prefixes because the retention factor is exponentiated by the prefix length.

\paragraph{Geometric approximation.}
For interpretation, suppose that the intrinsic predictability is approximately stable across positions:
\begin{equation}
\gamma_j(x)\approx \gamma_x,
\quad
0<\gamma_x<1.
\end{equation}
Define the effective survival factor
\begin{equation}
\rho_x(b)
=
\gamma_x r(b;B).
\end{equation}
Since $0<\gamma_x<1$ and $0<r(b;B)\le 1$, we have
\begin{equation}
0<\rho_x(b)<1.
\end{equation}
Then Eq.~\eqref{eq:tau_predictability_retention} reduces to
\begin{equation}
\tau(b;x)
\approx
\sum_{i=1}^{b}
\rho_x(b)^i.
\end{equation}
Using the finite geometric series identity, we obtain
\begin{equation}
\label{eq:tau_geometric}
\tau(b;x)
\approx
\frac{
\rho_x(b)
\left[
1-\rho_x(b)^b
\right]
}{
1-\rho_x(b)
}.
\end{equation}

Equation~\eqref{eq:tau_geometric} explains the mechanism behind sample-adaptive block-size selection. The factor $\gamma_x$ captures sample-level predictability: structured or deterministic inputs tend to have larger $\gamma_x$ and can sustain longer verified prefixes, whereas uncertain or open-ended inputs have smaller $\gamma_x$ and experience faster survival decay. Meanwhile, $r(b;B)$ captures block-size-induced proposal degradation and discourages choices far from the training configuration. Therefore, the optimal block size is governed by the interaction between sample predictability and block-size retention.

\paragraph{Local candidate interval.}
For a candidate set $\mathcal{B}$, the sample-wise optimal block size is defined as
\begin{equation}
B^*(x)
=
\arg\max_{b\in\mathcal{B}}
\tau(b;x).
\end{equation}
Under the retention model above, block sizes far from $B$ are penalized through $r(b;B)^i$, especially for longer prefixes. This provides a theoretical rationale for restricting the candidate space to a local interval around the training block size:
\begin{equation}
\mathcal{B}_{\mathrm{loc}}
=
\{b\in\mathcal{B}: |b-B|\le k\}.
\end{equation}
Empirically, this local interval captures the optimal block size for most samples. The resulting conclusion is that the best block size is sample-dependent, but it is expected to concentrate near the block size used to train the diffusion draft model. This locality reduces the adaptive block-size selection problem from an expensive global search to a structured local decision problem.

\subsection{Regret of Local Block-Size Prediction}

\paragraph{Acceptance-length regret.}
We finally analyze how prediction error affects the acceptance length. Let $B^*(x)$ be the optimal block size defined over the candidate set $\mathcal{B}$, and let $\hat{B}(x)\in\mathcal{B}$ be the block size predicted by the learned predictor. The acceptance-length regret is defined as
\begin{equation}
R(x)
=
\tau(B^*(x);x)
-
\tau(\hat{B}(x);x).
\end{equation}

Since block size is selected from a discrete candidate set, we assume a discrete Lipschitz condition over $\mathcal{B}$:
\begin{equation}
|\tau(b_1;x)-\tau(b_2;x)|
\le
L_\tau |b_1-b_2|,
\quad
\forall b_1,b_2\in\mathcal{B},
\end{equation}
for some constant $L_\tau>0$. Then
\begin{equation}
\begin{aligned}
R(x)
&=
\tau(B^*(x);x)
-
\tau(\hat{B}(x);x) \\
&\le
\left|
\tau(B^*(x);x)
-
\tau(\hat{B}(x);x)
\right| \\
&\le
L_\tau
|B^*(x)-\hat{B}(x)|.
\end{aligned}
\end{equation}
Taking expectation over the data distribution gives
\begin{equation}
\mathbb{E}[R(x)]
\le
L_\tau
\mathbb{E}
\left[
|B^*(x)-\hat{B}(x)|
\right].
\end{equation}

This bound shows that the loss in acceptance length is controlled by the distance between the predicted and optimal block sizes. Exact prediction is therefore not strictly necessary: if the predictor selects a block size close to the optimum, the regret remains bounded. This supports two design choices of our method. First, the prediction problem can be restricted to a local candidate set around $B$, where neighboring block sizes have similar acceptance behavior. Second, the predictor can be lightweight, since it does not need to solve a global optimization problem, but only needs to identify a near-optimal block size within a structured local interval.

\section{Performance Evaluation on Instruction and Code Models}
\label{app:main}
Table~\ref{tab:llama_qwen_speedup_acceptance} reports results on Llama-3.1-8B-Instruct and Qwen3-Coder-30B-A3B across Math, Code, and Chat benchmarks. Overall, our method consistently achieves the best performance across both models and decoding settings.

Under temperature $=0$, our method attains the highest average speedup on both Llama and Qwen, reaching 3.25$\times$ and 4.12$\times$, respectively, while also achieving the best or near-best average acceptance length $\tau$. In particular, on Qwen, it significantly outperforms all DFlash variants, improving speedup from 3.86$\times$ (DFlash(16)) to 4.12$\times$, with consistent gains in $\tau$. Under temperature $=1$, similar trends hold. Our method achieves 2.40$\times$ and 3.95$\times$ average speedups on Llama and Qwen, respectively, outperforming all baselines across most benchmarks. Notably, even under higher sampling uncertainty, it maintains competitive or superior acceptance lengths, indicating stable draft quality.

Across task categories, the improvements are consistent on Math, Code, and Chat benchmarks, suggesting that the benefits of adaptive block selection generalize across heterogeneous workloads and model scales. Overall, these results further confirm the effectiveness and robustness of the proposed method across both medium-scale and large-scale LLMs.

\begin{table*}[ht]
\centering
\scriptsize
\setlength{\tabcolsep}{3.0pt}
\renewcommand{\arraystretch}{1.14}
\caption{Speedup ratios and average acceptance length $\tau$ on Llama and Qwen models across Math, Code, and Chat benchmarks. Llama denotes Llama-3.1-8B-Instruct, and Qwen denotes Qwen3-Coder-30B-A3B. DFlash$(n)$ denotes DFlash with block size $n$.}
\label{tab:llama_qwen_speedup_acceptance}
\resizebox{\textwidth}{!}{%
\begin{tabular}{@{}ll*{8}{cc}@{}}
\toprule
\multirow{2}{*}{\textbf{Model}}
& \multirow{2}{*}{\textbf{Method}}
& \multicolumn{6}{c}{\textbf{Math}}
& \multicolumn{6}{c}{\textbf{Code}}
& \multicolumn{2}{c}{\textbf{Chat}}
& \multicolumn{2}{c}{\textbf{Overall}} \\
\cmidrule(lr){3-8}
\cmidrule(lr){9-14}
\cmidrule(lr){15-16}
\cmidrule(l){17-18}
&
& \multicolumn{2}{c}{GSM8K}
& \multicolumn{2}{c}{MATH-500}
& \multicolumn{2}{c}{AIME24}
& \multicolumn{2}{c}{HumanEval}
& \multicolumn{2}{c}{MBPP}
& \multicolumn{2}{c}{SWE-Bench}
& \multicolumn{2}{c}{MT-Bench}
& \multicolumn{2}{c}{Avg.} \\
\midrule
\multicolumn{2}{@{}l}{\textit{Temperature} $=0$}
& \textit{Speedup} & $\tau$
& \textit{Speedup} & $\tau$
& \textit{Speedup} & $\tau$
& \textit{Speedup} & $\tau$
& \textit{Speedup} & $\tau$
& \textit{Speedup} & $\tau$
& \textit{Speedup} & $\tau$
& \textit{Speedup} & $\tau$ \\
\midrule
\multirow{6}{*}{Llama}
& EAGLE-3     & 1.81$\times$ & 3.06 & 1.70$\times$ & 2.98 & 1.64$\times$ & 2.99 & 1.94$\times$ & 3.25 & 1.87$\times$ & 3.22 & 1.60$\times$ & 2.80 & 1.50$\times$ & 2.72 & 1.72$\times$ & 3.00 \\
& DFlash (4)  & 2.21$\times$ & 3.16 & 2.07$\times$ & 3.08 & 2.00$\times$ & 3.09 & 2.36$\times$ & 3.35 & 2.28$\times$ & 3.32 & 1.95$\times$ & 2.90 & 1.83$\times$ & 2.82 & 2.10$\times$ & 3.10 \\
& DFlash (8)  & 2.89$\times$ & 4.13 & 2.93$\times$ & 4.37 & 3.03$\times$ & 4.69 & 3.42$\times$ & 4.86 & 3.39$\times$ & 4.94 & 2.51$\times$ & 3.74 & 2.33$\times$ & 3.59 & 2.93$\times$ & 4.33 \\
& DFlash (16) & 2.91$\times$ & 4.16 & 3.11$\times$ & 4.63 & 2.93$\times$ & 4.54 & 3.67$\times$ & 5.21 & 3.57$\times$ & 5.20 & 2.52$\times$ & 3.76 & 2.52$\times$ & 3.88 & 3.03$\times$ & 4.48 \\
& DFlash (32) & 2.87$\times$ & 4.09 & 2.79$\times$ & 4.16 & 2.80$\times$ & 4.33 & 3.59$\times$ & 5.09 & 3.57$\times$ & 5.20 & 2.51$\times$ & 3.74 & 2.66$\times$ & 4.09 & 2.97$\times$ & 4.39 \\
& \textbf{BlockPilot} & \textbf{2.96$\times$} & \textbf{4.22} & \textbf{3.21$\times$} & \textbf{4.79} & \textbf{3.63$\times$} & \textbf{5.62} & \textbf{3.74$\times$} & \textbf{5.31} & \textbf{3.75$\times$} & \textbf{5.46} & \textbf{2.71$\times$} & \textbf{4.04} & \textbf{2.78$\times$} & \textbf{4.28} & \textbf{3.25$\times$} & \textbf{4.82} \\
\midrule
\multirow{6}{*}{Qwen}
& EAGLE-3     & 1.75$\times$ & 2.94 & 1.85$\times$ & 3.07 & 1.80$\times$ & 3.00 & 1.92$\times$ & 3.41 & 1.89$\times$ & 3.31 & 1.47$\times$ & 2.53 & 1.21$\times$ & 2.34 & 1.70$\times$ & 2.94 \\
& DFlash (4)  & 2.13$\times$ & 3.04 & 2.26$\times$ & 3.17 & 2.19$\times$ & 3.10 & 2.34$\times$ & 3.51 & 2.30$\times$ & 3.41 & 1.79$\times$ & 2.63 & 1.48$\times$ & 2.44 & 2.07$\times$ & 3.04 \\
& DFlash (8)  & 2.94$\times$ & 4.20 & 3.34$\times$ & 4.70 & 3.04$\times$ & 4.32 & 3.86$\times$ & 5.79 & 3.69$\times$ & 5.47 & 2.21$\times$ & 3.24 & 2.01$\times$ & 3.32 & 3.01$\times$ & 4.43 \\
& DFlash (16) & 3.58$\times$ & 5.12 & 4.16$\times$ & 5.85 & 3.67$\times$ & 5.20 & 5.74$\times$ & 8.61 & 5.11$\times$ & 7.58 & 2.48$\times$ & 3.65 & 2.29$\times$ & 3.77 & 3.86$\times$ & 5.68 \\
& DFlash (32) & 3.38$\times$ & 4.83 & 3.91$\times$ & 5.49 & 3.43$\times$ & 4.86 & 5.60$\times$ & 8.40 & 4.87$\times$ & 7.22 & 2.35$\times$ & 3.46 & 2.21$\times$ & 3.64 & 3.68$\times$ & 5.42 \\
& \textbf{BlockPilot} & \textbf{3.75$\times$} & \textbf{5.36} & \textbf{4.37$\times$} & \textbf{6.14} & \textbf{4.19$\times$} & \textbf{5.95} & \textbf{6.05$\times$} & \textbf{9.07} & \textbf{5.26$\times$} & \textbf{7.80} & \textbf{2.70$\times$} & \textbf{3.96} & \textbf{2.49$\times$} & \textbf{4.10} & \textbf{4.12$\times$} & \textbf{6.05} \\
\midrule
\multicolumn{2}{@{}l}{\textit{Temperature} $=1$}
& \textit{Speedup} & $\tau$
& \textit{Speedup} & $\tau$
& \textit{Speedup} & $\tau$
& \textit{Speedup} & $\tau$
& \textit{Speedup} & $\tau$
& \textit{Speedup} & $\tau$
& \textit{Speedup} & $\tau$
& \textit{Speedup} & $\tau$ \\
\midrule
\multirow{6}{*}{Llama}
& EAGLE-3     & 1.51$\times$ & 2.79 & 1.19$\times$ & 2.28 & 0.76$\times$ & 1.35 & 1.74$\times$ & 2.95 & 1.72$\times$ & 2.94 & 1.24$\times$ & 2.20 & 0.92$\times$ & 1.76 & 1.30$\times$ & 2.32 \\
& DFlash (4)  & 1.84$\times$ & 2.89 & 1.45$\times$ & 2.38 & 0.93$\times$ & 1.45 & 2.12$\times$ & 3.05 & 2.10$\times$ & 3.04 & 1.51$\times$ & 2.30 & 1.12$\times$ & 1.86 & 1.58$\times$ & 2.42 \\
& DFlash (8)  & 2.16$\times$ & 3.40 & 1.67$\times$ & 2.75 & 1.11$\times$ & 1.74 & 2.94$\times$ & 4.22 & 2.92$\times$ & 4.24 & 1.47$\times$ & 2.24 & 1.42$\times$ & 2.36 & 1.96$\times$ & 2.99 \\
& DFlash (16) & 2.45$\times$ & 3.85 & 1.75$\times$ & 2.88 & 1.15$\times$ & 1.80 & 3.00$\times$ & 4.31 & 2.80$\times$ & 4.06 & 1.60$\times$ & 2.44 & 1.54$\times$ & 2.56 & 2.04$\times$ & 3.13 \\
& DFlash (32) & 1.76$\times$ & 2.76 & 1.65$\times$ & 2.71 & 0.99$\times$ & 1.55 & 2.85$\times$ & 4.10 & 3.05$\times$ & 4.42 & 1.52$\times$ & 2.32 & 1.54$\times$ & 2.56 & 1.91$\times$ & 2.92 \\
& \textbf{BlockPilot} & \textbf{2.73$\times$} & \textbf{4.29} & \textbf{2.10$\times$} & \textbf{3.45} & \textbf{1.48$\times$} & \textbf{2.32} & \textbf{3.29$\times$} & \textbf{4.72} & \textbf{3.37$\times$} & \textbf{4.88} & \textbf{1.89$\times$} & \textbf{2.88} & \textbf{1.92$\times$} & \textbf{3.20} & \textbf{2.40$\times$} & \textbf{3.68} \\
\midrule
\multirow{6}{*}{Qwen}
& EAGLE-3     & 1.69$\times$ & 2.83 & 1.71$\times$ & 2.96 & 1.59$\times$ & 2.74 & 1.91$\times$ & 3.34 & 1.82$\times$ & 3.20 & 1.34$\times$ & 2.27 & 1.20$\times$ & 2.27 & 1.61$\times$ & 2.80 \\
& DFlash (4)  & 2.06$\times$ & 2.93 & 2.09$\times$ & 3.06 & 1.94$\times$ & 2.84 & 2.33$\times$ & 3.44 & 2.22$\times$ & 3.30 & 1.63$\times$ & 2.37 & 1.46$\times$ & 2.37 & 1.96$\times$ & 2.90 \\
& DFlash (8)  & 2.89$\times$ & 4.10 & 3.05$\times$ & 4.47 & 2.56$\times$ & 3.76 & 3.77$\times$ & 5.58 & 3.67$\times$ & 5.44 & 1.96$\times$ & 2.86 & 2.01$\times$ & 3.28 & 2.84$\times$ & 4.21 \\
& DFlash (16) & 3.41$\times$ & 4.84 & 3.72$\times$ & 5.45 & 3.09$\times$ & 4.54 & 5.27$\times$ & 7.79 & 4.91$\times$ & 7.28 & 2.09$\times$ & 3.04 & 2.27$\times$ & 3.70 & 3.54$\times$ & 5.23 \\
& DFlash (32) & 3.41$\times$ & 4.84 & 3.61$\times$ & 5.29 & 2.88$\times$ & 4.22 & 5.54$\times$ & 8.19 & 4.89$\times$ & 7.25 & 2.09$\times$ & 3.05 & 2.08$\times$ & 3.38 & 3.50$\times$ & 5.17 \\
& \textbf{BlockPilot} & \textbf{3.77$\times$} & \textbf{5.36} & \textbf{4.01$\times$} & \textbf{5.87} & \textbf{3.33$\times$} & \textbf{4.89} & \textbf{6.35$\times$} & \textbf{9.39} & \textbf{5.34$\times$} & \textbf{7.92} & \textbf{2.36$\times$} & \textbf{3.44} & \textbf{2.48$\times$} & \textbf{4.04} & \textbf{3.95$\times$} & \textbf{5.84} \\
\bottomrule
\end{tabular}%
}
\end{table*}

\section{Limitations and Future Work}
\label{Limitations}
In this paper, our data construction method provides an effective way to identify suitable block sizes for different training samples, enabling the model to learn more fine-grained execution preferences. However, this benefit comes with some computational burden, particularly for very large models. For example, for a 32B model, executing a single sample under one block size takes approximately 5 seconds. To construct one training sample, we evaluate all candidate block sizes in the range $\{B-k,\dots,B+k\}$. Under our default setting of $k=2$, this requires five executions with different block sizes, resulting in roughly 25 seconds per training sample. Nevertheless, this overhead is incurred only once during data construction and can be performed entirely offline, without affecting the efficiency of model training or inference. Although this cost is acceptable in our setting, it could be further reduced by more efficient search strategies. Since optimizing the data construction pipeline is not the main focus of this work, we leave this direction to future work. Possible improvements include heuristic search strategies, adaptive block-size pruning, and early stopping mechanisms to avoid evaluating unnecessary candidates. In addition, one could explore lightweight proxy metrics to estimate the effectiveness of different block sizes before full execution, or adopt a coarse-to-fine search strategy that first evaluates a small set of representative candidates and then refines the search around promising regions. Another possible direction is to reuse execution results across similar samples, thereby reducing redundant evaluations during data construction.

\newpage

\end{document}